\documentclass[12pt]{article}
\usepackage{amsmath}
\usepackage{graphicx}
\usepackage{enumerate}
\usepackage{natbib}

\usepackage[utf8]{inputenc} 
\usepackage[T1]{fontenc}	
\usepackage{lmodern}    
\usepackage{hyperref}       
\usepackage{url}            
\usepackage{booktabs}       
\usepackage{amsfonts}       
\usepackage{nicefrac}       
\usepackage{microtype}      
\usepackage{comment}

\usepackage{bbm}
\usepackage{wrapfig}
\usepackage{bm}
\usepackage{mathtools}
\usepackage{amsthm}
\usepackage{array}
\DeclareMathOperator*{\argmax}{arg\,max}
\DeclareMathOperator*{\argmin}{arg\,min}
\usepackage[english]{babel}
\usepackage[ruled,vlined]{algorithm2e}
\SetKwProg{Init}{initialize}{}{}
\usepackage{amssymb}
\usepackage{color,soul}
\usepackage[colorinlistoftodos,prependcaption,textsize=scriptsize]{todonotes}
\usepackage{xcolor}
\DeclarePairedDelimiterX{\infdivx}[2]{(}{)}{%
  #1\;\delimsize\|\;#2%
}

\newcommand{\blind}{0}

\begin{document}

\def\spacingset#1{\renewcommand{\baselinestretch}%
{#1}\small\normalsize} \spacingset{1}


\if0\blind
{
  \title{\bf A Continual Learning Framework for Adaptive Defect Classification and Inspection}
  \author{Wenbo Sun\\
    University of Michigan Transportation Research Institute\\
    and \\
    Raed Al Kontar \\
    Industrial and Operations Engineering, University of Michigan, Ann Arbor\\
    and \\
    Judy Jin \\
    Industrial and Operations Engineering, University of Michigan, Ann Arbor\\
    and \\
    Tzyy-Shuh Chang \\
    OG Technologies, Inc.}
  \maketitle
} \fi

\if1\blind
{
  \bigskip
  \bigskip
  \bigskip
  \begin{center}
    {\LARGE\bf A Continual Learning Framework for Adaptive Defect Classification and Inspection}
\end{center}
  \medskip
} \fi

\bigskip
\begin{abstract}
Machine-vision-based defect classification techniques have been widely adopted for automatic quality inspection in manufacturing processes. This article describes a general framework for classifying defects from high volume data batches with efficient inspection of unlabelled samples. The concept is to construct a detector to identify new defect types, send them to the inspection station for labelling, and dynamically update the classifier in an efficient manner that reduces both storage and computational needs imposed by data samples of previously observed batches. Both a simulation study on image classification and a case study on surface defect detection via 3D point clouds are performed to demonstrate the effectiveness of the proposed method.
\end{abstract}

\noindent%
\vfill

\newpage
\spacingset{1.5} 

\section{Introduction}

Recent development of advanced sensing and high computing technologies has enabled the wide adoption of machine vision to automatically inspect products’ dimensional quality for efficient process control and reducing the manual inspection cost. The process control procedure requires effective data analysis methods to provide reliable inspection results. In this paper, we consider a high-volume manufacturing system that uses machine vision at the quality inspection station for automatic classification of product defects. Here classification implies both; identifying a defect and classifying its corresponding type.  

\begin{figure}
    \centering
    \includegraphics[scale=0.43]{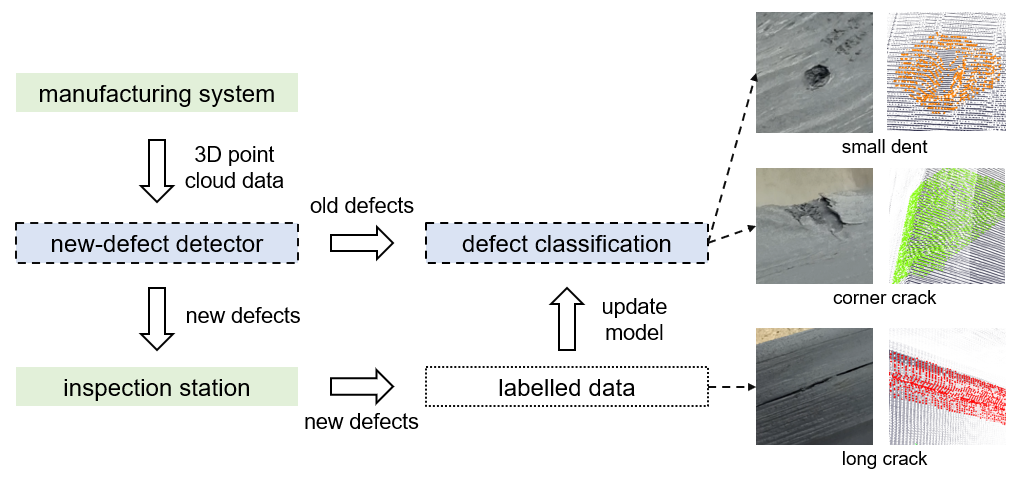}
    \caption{\it{Illustration of the proposed statistical framework on adaptive defect classification of 3D point cloud data. The dashed boxes highlight the two key modules in the proposed framework.}}
    \label{fig:3dpt}
\end{figure}

As a motivating example, we consider the scenario where batches of three-dimensional (3D) point cloud data are independently collected from a manufacturing process. The 3D point cloud data is obtained by measuring the 3D location of points on the product surface using a 3D scanner. The location measurements can then be used for fast classification of surface defects, and thus provide timely feedback for process control. Fig.~\ref{fig:3dpt} (right) shows some exemplar surface defects on a wood product and the corresponding 3D point cloud measurements. 

The 3D point cloud measurements have a set of defining characteristics that should be considered in the development of defect classification techniques. (i) The data size of 3D point cloud measurements is very high. In the aforementioned example, a single wood part of length $2.85$ meters incurs a data size of $82.7$ megabytes. These parts are inspected sequentially in batches. As a result, defect classification techniques should be able to learn from both old and current batches while accounting for limited storage and computational capabilities needed to achieve continuous process monitoring. (ii) Product defects have various types/classes that vary in shape and size and can occur at different locations. As a result, it is infeasible to pre-collect all types of defects as training samples. Furthermore, it is very common to have new defect types that evolve with time. Therefore, it is critical to continuously learn new defect types and augment the classification schemes with the ability to classify them. (iii) Manual data labeling for 3D point cloud measurements is extremely time-consuming and tedious as it requires identifying 3D locations of all individual scanned points within the range of defects. Therefore, it is desirable to only request manual inspection and data labeling for new defect types while avoiding or reducing data labeling for previously well-trained old defects. 

Indeed, recent classification techniques have been developed for defect detection \citep{jovanvcevic20173d, hackel2017joint} or object recognition \citep{qi2017pointnet} based on 3D point cloud data. However, they require one to pre-label all the 3D point cloud data before training the classification model, which is infeasible for a high-volume manufacturing system and makes it difficult to account for new defect types that are either missed at initial labeling or evolve with time. In this paper, we take an alternative route through continual learning of a defect classifier. Specifically, we present a statistical framework that can automatically detect new defect types, which are sent to the inspection station, and dynamically update the classifier in an efficient manner that reduces both storage and computational needs imposed by data samples of previously observed batches. Although machine vision is widely used for quality inspection in many manufacturing processes, it is often the case that machine vision may not always provide accurate inspection results, thus a double-check inspection station is built to further confirm whether the alarmed products are truly defective. In this paper, we assume that the inspection station provides $100\%$ correct labelling results for the alarmed samples but those inspections are time consuming or costly. Our idea and framework is illustrated in Fig.~\ref{fig:3dpt}. Figure~\ref{fig:3dpt} shows the flowchart of our proposed approach for inspecting a current batch having two old defect types (small dent \& corner crack) that have been seen in the previous batches, and one new defect type (long crack) that occurs for the first time. Our approach is consisted of two modules. One is a new-defect detector that can automatically detect old defect types and separate them from new ones. In turn, efforts for data labelling of old defect types can be reduced. The other module is an updated defect classifier that gives the new inspection results. This updated classifier is intended to efficiently update the old model while relaxing the storage and computational needs for old defect types.  

Regarding the two modules. The first module separates new defect types from old ones. An intuitive way is to monitor the probabilities or likelihoods of the current batch data falling within the previously trained defects. This will generate an alarm for new defect types when all the probabilities are low. However, state-of-the-art classifiers often over-estimate such probabilities, and thus over-confidently classify a sample from a new defect type into an old defect type \citep{nguyen2015deep}. To resolve this issue, one may exploit recent out-of-distribution detection techniques. The basic idea is to construct a score that separates in-distribution samples (here samples from old defect types) from out-of-distribution samples (here samples from new defect types). Such score functions include but are not limited to the ODIN score \citep{liang2017enhancing}, Mahalanomis-distance-based score \citep{lee2018simple, hsu2020generalized}, energy-based score \citep{liu2020energy}, and feature space singularity distance \citep{huang2020feature}. These score functions are often parametrized by some tuning parameters, whose values are often trained by optimally separating the samples in the new classes from those in the old classes. However, a critical gap of applying the out-of-distribution technique in the proposed framework is that the training samples from the new defect types are not available, leading to a poor new-defect detection performance.

The second module is to classify defect types via a classification model. With the accumulated samples of newly labelled defect types, the model for defect classification should be continuously updated. Under constrained computational resources, only limited samples from the old defect types and the old model parameters can be stored and used to update the model. This is accomplished via continual learning techniques. The idea of continual learning dates back decades ago, to mixed-effects modeling where instead of re-learning a new model, empirical Bayes is used to updated random effects conditioned on the new observed data. One of the seminal works in this area is \citep{gebraeel2005residual} who tests the developed model on bearing degradation signals. In contrast, our work focuses on classification, rather than regression tasks. Some works in continual learning of classification models include, \cite{syed1999incremental}, who updates a support vector machine in a batch learning mode, and \cite{ozawa2008incremental}, who extends incremental principal component analysis to classify chunks of training samples. To deal with high volume and high dimensional dataset, deep-neural-network-based continual learning techniques have been developed, which build deep neural networks that can learn new classes while retaining the classification accuracy of old classes. For example, \cite{xiao2014error} and \cite{roy2020tree} propose two hierarchical architectures to allow deep neural networks grow sequentially when new classes of samples arrive. \cite{kochurov2018bayesian}, \cite{kirkpatrick2017overcoming} and \cite{li2017learning} impose regularization terms on the loss function to maintain classification accuracy for old classes. \cite{rebuffi2017icarl} stores representative samples from the old classes and uses them to update the model. \cite{rusu2016progressive} and \cite{mallya2018packnet} introduce additional model components to draw the connection of model parameters between the old and new classes. \cite{wu2019large} and \cite{he2020incremental} propose methods to update the model in an online manner when the old and new classes arrive simultaneously, and investigate the data imbalance issue between the old and new classes. However, these continual learning techniques rely on fully labelled samples in both of the old and new classes, which is not available in the target manufacturing application. Moreover, without additional treatments to the existing continual learning methods, the learnt classifier tends to over-confidently classify samples from incoming new classes into the classes in the training dataset.

To overcome the shortcomings of the existing methods, this paper presents an adaptive defect classification approach that integrates both out-of-distribution learning and continual learning together. The key contribution is in learning a ''Look-ahead" defect classifier that aims to separate the known defects and potential unknown new defects in future data batches. Indeed, the engineering setting of having an "double-check" inspection station to confirm machine-vision results is the key motivation behind this new framework. To the best of our knowledge, the proposed approach is the first anomaly detection work under this new engineering setting, which is highly demanded in manufacturing industry for automatic quality inspection using a machine vision system.

In particular, the contributions of the paper are three-folds. (a) Going beyond the existing continual learning methods, we propose a data-driven strategy for essential data labelling that is required only for new defect types; (b) Different from the existing out-of-distribution learning techniques, we develop a systematic way to train a classifier that simultaneously updates and learns new defect types from inspections and avoids over-confident incorrect classification of samples from unseen classes. Our approach does not require all samples from previous batches to be stored or re-trained. (c) provide a guideline for selecting auxiliary out-of-distribution samples in different engineering applications. The proposed method is generic and compatible with any type of classifier and out-of-distribution score function. All these demonstrate important new contributions and the engineering significance of the proposed method.

The rest of paper is organized as follows. In Section~\ref{s:notation}, we will describe the notations, data structure, and objective. In Section~\ref{s:method}, the proposed method is elaborated for adaptive defect classification. Section~\ref{s:simulation} and Section~\ref{s:case} demonstrate the effectiveness of the method by using a simulation study in image classification and a case study in surface defect classification for 3D point clouds, respectively. The paper then ends with a conclusion in Section~\ref{s:conclusion}.

\section{Notations, data structure and objective}\label{s:notation}
In this paper, we consider a high-volume discrete manufacturing process that has a machine vision system to automatically detect potential defects and a double-check inspection station is followed to confirm whether those alarmed parts are truly detective. It should be clarified that those alarmed samples by the machine vision system will be correctly labeled in the inspection station, while those un-alarmed samples will not go through the inspection station, thus no labelling.
During production, the products are discretely produced and the inspection results come in batches. In each batch, the test results are assumed as independent samples. We call this data setting as discretized data batches, which is commonly seen in many manufacturing applications. In this situation, the inspection results are considered as independent samples.

The space of measurement data is denoted by $\mathcal{X}$. \{Suppose independent measurement samples arrive in multiple batches $\left\{0,\dots,T\right\}$ from a discrete manufacturing process. Let $\boldsymbol{X}^{(t)}=\left\{\boldsymbol{x}_1^{(t)},\dots,\boldsymbol{x}_{n^{(t)}}^{(t)}\right\}$ denote the set of $n^{(t)}$ measurement data at batch $t\in\left\{0,\dots,T\right\}$, where $\boldsymbol{x}_i^{(t)}\in\mathcal{X}$ is the $i$-th sample in $X^{(t)}$. In general, $\boldsymbol{x}_i^{(t)}$ can be any type of measurement data that are commonly seen in manufacturing processes, such as profile signals, images, time-series, etc.. Given that the measurement data can take any form, such as vectors, matrices or tensors, we denote vectors, matrices or tensors by using lowercase boldface letters, and denote a set of vectors, matrices or tensors by using uppercase boldface letters throughout the paper. Let $y_i^{(t)}$ denote the class label corresponding to $\boldsymbol{x}_i^{(t)}$. In particular, $y_i^{(t)}=0$ when sample $i$ is non-defective, and $y_i^{(t)}$ takes a positive integer value corresponding to the defect class when the sample $i$ is defective. Our major objective is to predict the quality label $y_i^{(t)}$ via a classifier $f:\mathcal{X}\rightarrow \mathbb{R}$ for defect classification.

Let $\mathcal{Y}^{(t)}$ denote the set of all the class labels up to batch $t$. When the defect types come in a class-incremental form, i.e., $\mathcal{Y}^{(t)}\not\subset\mathcal{Y}^{(t-1)}$, the classifier $f$ trained at the previous batches cannot predict the newly appeared defect types from $\mathcal{Y}^{(t)} \backslash \mathcal{Y}^{(t-1)}$. To classify these new defects, it requires to update the classifier $f$ with labelled samples from a subset of $\boldsymbol{X}^{(t)}$. Given that inspecting all the $\boldsymbol{X}^{(t)}$'s class labels are expensive, we would like to focus on the samples whose labels belong to the set $\mathcal{Y}^{(t)}\not\subset\mathcal{Y}^{(t-1)}$. To do this, a new-defect detector $g:\mathcal{X}\rightarrow\left\{0,1\right\}$ is developed to screen the newly appeared defects. Specifically, we would like $g\left(\boldsymbol{x}_i^{(t)}\right)$ to return $1$ if $y_i^{(t)}\in \mathcal{Y}^{(t)} \backslash \mathcal{Y}^{(t-1)}$, and to return $0$, otherwise. 

\begin{figure}
    \centering
    \includegraphics[height=4.2in]{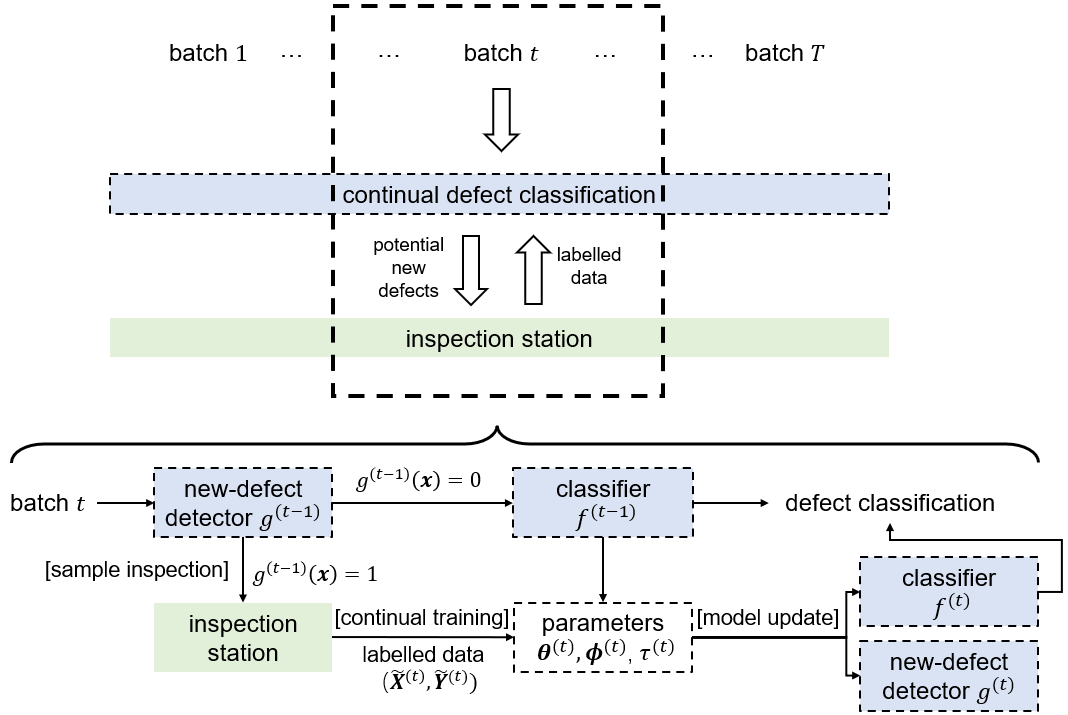}
    \caption{\it{Flowchart of the proposed statistical framework for data batch $t$. The top panel (above the brace) illustrates the overall methodology flowchart among all the data batches. The bottom panel (below the brace) depicts the detailed information flow within each data batch. The shaded dashed boxes highlight the key modules. The key functions of the proposed algorithm is highlighted in the square brackets. After model update, the classifier $f^{(t)}$ can be used to classify all the defect types appeared in data batches $1,\dots,t$.}}
    \label{fig:flowchart}
\end{figure}

Now we add the superscript $(t-1)$ to the classifier $f$ and the new-defect detector $g$ to denote the trained functions prior to batch $t$. The proposed statistical framework for adaptive defect classification works as follows. At batch $t=0$, $f^{(0)}$ and $g^{(0)}$ are initially trained using historical dataset. At batch $t>0$, $g^{(t-1)}$ is firstly applied to $\boldsymbol{X}^{(t)}$ to detect potential new defect types, that is, to select $\boldsymbol{\widetilde{X}}^{(t)}=\left\{\boldsymbol{x}:g^{(t-1)}(\boldsymbol{x})=1, \boldsymbol{x}\in \boldsymbol{X}^{(t)}\right\}$. The samples of $\widetilde{\boldsymbol{X}}^{(t)}$ are then sent to the inspection station for labelling, and denoted by $\boldsymbol{\widetilde{Y}}^{(t)}$. These labelled samples are further used to update the classifier $f^{(t)}$ and the new-defect detector $g^{(t)}$. The procedure is repeatedly applied until the last batch $T$ is processed. A flowchart of the above procedure is provided in Fig.~\ref{fig:flowchart}. The proposed statistical framework is well established once the updating strategy of $f^{(t)}$ and $g^{(t)}$ is determined. In the next section, we will elaborate the technical details when updating $f^{(t)}$ and $g^{(t)}$.

\section{Method}\label{s:method}
In this section, we decompose the proposed algorithm into three sub-functions and explain how they are used for adaptive defect classification. Section~\ref{ss:overall} provides an overall structure of the proposed algorithms. Section~\ref{ss:inspection} to \ref{ss:detector} elaborate the algorithmic details in three functions. Section~\ref{ss:practical} discusses the practical considerations including the choices of models, auxiliary out-of-distribution samples and the processes for tuning parameters.

\subsection{Adaptive defect classification}\label{ss:overall}
To begin with, we specify the definitions of the classifier and new-defect detector as follows. Let $\left|\mathcal{Y}^{(t)}\right|$ denote the total number of labelled classes until batch $t$. The classification results depend on the predicted probabilities of a sample falling in each class $j$, which is denoted by $q_j\left(\cdot|\theta^{(t)}\right):\mathcal{X}\rightarrow \mathbb{R}$, with $\boldsymbol{\theta}^{(t)}$ representing the model parameter of the classifier. Let $\boldsymbol{q}$ be the vector that collects all the $q_j$'s, for $j=\left\{1,...,\left|\mathcal{Y}^{(t)}\right|\right\}$. Note that $\boldsymbol{q}$ can be the predicted probability function in any classifier. The classifier $f^{(t)}$ returns the class label as the maximal element in the vector $\boldsymbol{q}$, which is
\begin{equation}
    f^{(t)}(\boldsymbol{x})\equiv \argmax_j\,q_j\left(\boldsymbol{x}|\boldsymbol{\theta}^{(t)}\right).
    \label{eq:f}
\end{equation}

The new-defect detector is established based on a score function that evaluates the probability whether a sample comes from a new defect or not. Let $s\left(\cdot|\boldsymbol{\phi}^{(t)}\right)$ denote the score function, where $\boldsymbol{\phi}^{(t)}$ is the model parameter of the score function. $s$ maps the predicted probability function $\boldsymbol{q}$ to a real-valued score such that the scores of new defects are separable from those of old defects. The parametric form of $s$ can be flexibly selected from any out-of-distribution score functions. For example, the Mahalanobis-distance-based score measures the Mahalanobis distance between a sample $\boldsymbol{x}$ and the closest class-conditional Gaussian distribution. In particular, let $\boldsymbol{\hat\mu}_j$ and $\boldsymbol{\widehat\Sigma}$ denote the empirical mean and covariance matrix of $\boldsymbol{q}(\boldsymbol{x})$ for samples of defect type $j$, respectively, i.e.,
\begin{eqnarray}
    \boldsymbol{\hat\mu}_j &=& \frac{1}{N_j}\sum_{j:y_i=j}\boldsymbol{q}(\boldsymbol{x}_i),\nonumber\\
    \boldsymbol{\widehat\Sigma} &=& \frac{1}{N_j}\sum_j \sum_{i:y_i=j}\left(\boldsymbol{q}(\boldsymbol{x}_i)-\boldsymbol{\hat\mu}_j\right)\left(\boldsymbol{q}(\boldsymbol{x}_i)-\boldsymbol{\hat\mu}_j\right)^\text{T},
\end{eqnarray}
where $N_j$ is the number of training samples with defect type $j$. The Mahalanobis-distance-based score is then formally expressed as the \textbf{negative} distance from $\boldsymbol{q}\left(\boldsymbol{x}|\boldsymbol{\theta}^{(t)}\right)$ to the empirical mean of the softmax scores $\boldsymbol{\hat\mu}_j$ with considering the empirical covariance $\boldsymbol{\widehat\Sigma}$.
\begin{equation}
    s\left(\boldsymbol{q}\left(\boldsymbol{x}|\boldsymbol{\theta}^{(t)}\right)\boldsymbol{\phi}^{(t)}\right)=\max_j\,-\left(\boldsymbol{q}\left(\boldsymbol{x}|\boldsymbol{\theta}^{(t)}\right)-\boldsymbol{\hat\mu}_j\right)^\text{T}\,\boldsymbol{\widehat\Sigma}^{-1} \,\left(\boldsymbol{q}\left(\boldsymbol{x}|\boldsymbol{\theta}^{(t)}\right)-\boldsymbol{\hat\mu}_j\right).
\end{equation}

In this way, a low $s\left(\boldsymbol{q}\left(\boldsymbol{x}|\boldsymbol{\theta}^{(t)}\right)\boldsymbol{\phi}^{(t)}\right)$ implies that $\boldsymbol{q}\left(\boldsymbol{x}|\boldsymbol{\theta}^{(t)}\right)$ is far from the center of the softmax scores of old samples. Therefore, we set an upper threshold $\tau^{(t)}$ for $s\left(\boldsymbol{q}\left(\boldsymbol{x}|\boldsymbol{\theta}^{(t)}\right)\boldsymbol{\phi}^{(t)}\right)$, and report detection results when $s\left(\boldsymbol{q}\left(\boldsymbol{x}|\boldsymbol{\theta}^{(t)}\right)\boldsymbol{\phi}^{(t)}\right)$ is above the threshold. That is, to define the new-defect detector $g^{(t)}$ as:
\begin{equation}
    g^{(t)}\left(\boldsymbol{x}\right)\equiv\mathbb{I}\left\{s\left(\boldsymbol{q}\left(\boldsymbol{x}|\boldsymbol{\theta}^{(t)}\right)\big|\boldsymbol{\phi}^{(t)}\right)\leq\tau^{(t)}\right\},
    \label{eq:g}
\end{equation}
where $\mathbb{I}$ is the indicator function. Eq.(\ref{eq:g}) shows that the new-defect detector $g^{(t)}$ depends on the classifier's prediction $\boldsymbol{q}$, implying that the new-defect detector should be updated with the classifier among different data batches.

As is shown in the following Algorithm~\ref{a:overall}, the proposed method can be decomposed into three sub-functions to incrementally update information from the measurement data $\boldsymbol{X}^{(t)}$ at batch $t$. First, the previously trained new-defect detector $g^{(t-1)}$ is used to detect the measurement data of new defect types in $\boldsymbol{X}^{(t)}$. The inspection station labels these samples, combines the labelled data with a given number of data of old defect types, and returns the combined dataset as $\boldsymbol{D}^{(t)}$. Second, the combined dataset $\boldsymbol{D}^{(t)}$ is fed to a continual learner to update the parameters of the classifier $\boldsymbol{\theta}^{(t)}$, the parameter of the new-defect detector $\boldsymbol{\phi}^{(t)}$, and the threshold of the score function $\tau^{(t)}$. Lastly, the classifier and new-defect detector are updated according to Eq.(\ref{eq:f}) and Eq.(\ref{eq:g}), respectively. At the end of each batch $t$, $f^{(t)}$ learns the new defect types in $\boldsymbol{D}^{(t)}$, and $g^{(t)}$ will identify the new defect types beyond $\mathcal{Y}^{(t)}$. In the following subsections, we will elaborate the technical details in three sub-functions ``Sample Inspection'', ``Continual Training'', and ``Model Update''.

\begin{algorithm}[H]
    \SetAlgoLined
    \SetKwInOut{Input}{input}
    \SetKwInOut{Output}{output}
    
    \Input{a set of measurement data $\left\{\boldsymbol{X}^{(t)};t=0,\dots,T\right\}$, an initial set of class labels $\mathcal{Y}^{(0)}$, a pre-trained classifier $f^{(0)}$, an pre-trained new-defect detector $g^{(0)}$.}
    
    \For{$t \in \left\{1,...,T\right\}$}{
        $\boldsymbol{D}^{(t)},\mathcal{Y}^{(t)}\gets\text{Sample Inspection}\left(g^{(t-1)},\boldsymbol{X}^{(t)},\mathcal{Y}^{(t-1)},\left\{\boldsymbol{D}^{(k)},k=0,...,t-1\right\}\right)$
        $\boldsymbol{\theta}^{(t)},\boldsymbol{\phi}^{(t)},\tau^{(t)}\gets\text{Continual Training}\left(\left\{\boldsymbol{D}^{(k)},k=0,...,t\right\},\boldsymbol{\theta}^{(t-1)},\boldsymbol{\phi}^{(t-1)},\tau^{(t-1)}\right)$
        $f^{(t)},g^{(t)}\gets\text{Model Update}\left(\boldsymbol{\theta}^{(t)},\boldsymbol{\phi}^{(t)},\tau^{(t)}\right)$
    }
    
    \Output{the updated classifier $f^{(T)}$ and new-defect detector $g^{(T)}$}
    \caption{Continual anomaly detection}
    \label{a:overall}
\end{algorithm}

\subsection{Sample inspection}\label{ss:inspection}
At each batch $t$, the first step is let all the unlabelled measurement data $\boldsymbol{X}^{(t)}$ go through the new-defect detector to identify the samples of potential new defects, and then send them to the inspection station for labelling. Following the definition of the new-defect detector, we store the samples in $\boldsymbol{\widetilde{X}}^{(t)}=\left\{\boldsymbol{x}:g^{(t-1)}(\boldsymbol{x})=1\right\}$ and send them to the inspection station to get the corresponding labels $\boldsymbol{\widetilde{Y}}^{(t)}$. During inspection, $\mathcal{Y}^{(t-1)}$ is continuously updated to include all the new defect types that have not been seen in the previous batches, resulting in the updated set $\mathcal{Y}^{(t)}$. The dataset $\boldsymbol{D}^{(t)}$ then combines the newly labelled samples with the old samples in the previous batches, which will be further used to update the classifier and new-defect detector in the next step. Here we recommend to keep a given number of samples in the previous data batches because it can improve the classifier and new-defect detector's performance after updating the models. The determination of the number of samples in the previous data batches will be discussed in Section~\ref{ss:practical}. Let $\boldsymbol{D}_{old}^{(t)}$ denote the dataset of old samples we keep in the previous data batches. The proposed algorithm for sample inspection is displayed in Algorithm~\ref{a:inspection}.

It is worth noting that the performance of $g^{(t-1)}$ affects the efficiency of defect labelling and model update. In particular, for a sample $(\boldsymbol{x},y)$ with $g^{(t-1)}(\boldsymbol{x})=1$ and $y\in\mathcal{Y}^{(t-1)}$ (i.e. false alarm), the inspection station labels an old defect type. Such a labelled sample $(\boldsymbol{x},y)$ has a small contribution to the improvement of the classifier and new-defect detector's performance, making the inspection less efficient. For a sample $(\boldsymbol{x},y)$ satisfying $g^{(t-1)}(\boldsymbol{x})=0$ and $y\notin\mathcal{Y}^{(t-1)}$ (i.e. misdetection), the sample of new defect type $(\boldsymbol{x},y)$ remains unlabelled, and thus is not used for model update. An overly-high misdetection rate leads to a small sample size of new classes, and may hence result in a poor classification performance on new defects. In practice, unlike traditional statistical process control techniques, the misdetection rate is not necessarily near perfect, and should be determined based on the requirement on the sample size of each new class. For instance, suppose $10,000$ samples under a new defect type arrive in a batch, a new-defect detector with $50\%$ misdetection rate can still send $5,000$ samples for labelling, which is typically essential to train a classifier that learns the new defects.

\begin{algorithm}[H]
    \SetAlgoLined
    \SetKwInOut{Input}{input}
    \SetKwInOut{Output}{output}
    \Input{the previous detector $g^{(t-1)}$, the new data batch $\boldsymbol{X}^{(t)}$, the previous set of labels $\mathcal{Y}^{(t-1)}$, the previous batches $\left\{\boldsymbol{D}^{(k)},k=0,...,t-1\right\}$}
    \Init{}{$\boldsymbol{\widetilde{X}}^{(t)}=\left\{\right\}$\\
        $\boldsymbol{\widetilde{Y}}^{(t)}=\left\{\right\}$\\
        $\mathcal{Y}^{(t)}=\mathcal{Y}^{(t-1)}$\\}
    \For{$\boldsymbol{x}\in\mathcal{X}^{(t)}$}{
        \If{$g^{(t-1)}(\boldsymbol{x})==1$}{
        get label $y$ of sample $\boldsymbol{x}$ from the inspection station\\
        $\boldsymbol{\widetilde{X}}^{(t)} \gets \boldsymbol{\widetilde{X}}^{(t)} \bigcup \boldsymbol{x}$\\
        $\boldsymbol{\widetilde{Y}}^{(t)} \gets \boldsymbol{\widetilde{Y}}^{(t)} \bigcup y$\\
        \If{$y \not\in\mathcal{Y}^{(t)}$} { $\mathcal{Y}^{(t)} \gets \mathcal{Y}^{(t)} \bigcup y$}
        }
    }
    $\boldsymbol{D}^{(t)} \gets \left(\boldsymbol{\widetilde{X}}^{(t)},\boldsymbol{\widetilde{Y}}^{(t)}\right) \bigcup \boldsymbol{D}_{old}^{(t)}$\\
    \Output{Dataset with labelled new defects and updated set of known labels: $\boldsymbol{D}^{(t)},\mathcal{Y}^{(t)}$}
    \caption{Sample Inspection}
    \label{a:inspection}
\end{algorithm}

\subsection{Continual training}\label{ss:classifier}
In this sub-function, the classifier and new-defect detector are simultaneously updated based on the labelled dataset $\boldsymbol{D}^{(t)}$ and an auxiliary out-of-distribution dataset $\boldsymbol{\widetilde{D}}^{(t)}$. The loss function for model training is proposed as follows:
\begin{eqnarray}
    \left(\boldsymbol{\theta}^{(t)},\boldsymbol{\phi}^{(t)}\right)&=&\argmin_{\boldsymbol{\theta},\boldsymbol{\phi}}\,L\left(\boldsymbol{\theta},\boldsymbol{\phi},\tau^{(t-1)}\right)\nonumber\\
    L\left(\boldsymbol{\theta},\boldsymbol{\phi},\tau^{(t-1)}\right)&=&L_{cont}\left(\boldsymbol{\theta}\right)+\lambda_{ood}\sum_{(\boldsymbol{x}_{in},y_{in})\in \boldsymbol{D}^{(t)}} \max\left\{0,\tau^{(t-1)}-s\left(\boldsymbol{q}\left(\boldsymbol{x}_{in}|\boldsymbol{\theta}\right)\big|\boldsymbol{\phi}\right)\right\}\nonumber\\
    &+&\lambda_{ood}\sum_{(\boldsymbol{x}_{out},y_{out})\in \boldsymbol{\widetilde{D}}^{(t)}} \max\left\{0,-\tau^{(t-1)}+s\left(\boldsymbol{q}\left(\boldsymbol{x}_{out}|\boldsymbol{\theta}\right)\big|\boldsymbol{\phi}\right)\right\},
    \label{eq:opt}
\end{eqnarray}
where $L_{cont}\left(\boldsymbol{\theta}\right)$ represents a loss function in continual classification to be discussed shortly. The second and third terms in Eq.\ref{eq:opt}, which encourages the score function $s$ to be greater than the threshold $\tau^{(t-1)}$ for any known defects from $\boldsymbol{D}^{(t)}$, and to be smaller than the threshold $\tau^{(t-1)}$ for any auxiliary out-of-distribution samples from $\boldsymbol{\widetilde{D}}^{(t)}$. In this way, the trained classifier not only learns the new classes in $\boldsymbol{D}^{(t)}$, \textbf{but also separates the known defects and potential new defects in future data batches.} This indeed is reminiscent of the slack term in support vector machine \citep{suykens1999least} which aims to achieve an optimal separation across classes. 

Now regarding $L_{cont}\left(\theta\right)$, we propose to penalize the cross-entropy classification loss $CE$ with a penalty term on the normalized Gaussian distance between the updated and original model parameters. This is given as 
\begin{eqnarray}
    L_{cont}\left(\boldsymbol{\theta}\right)&=&L_{class}\left(\boldsymbol{\theta}\right)+\lambda_{prior} L_{prior}\left(\boldsymbol{\theta}\right),\nonumber\\
    L_{class}\left(\boldsymbol{\theta}\right)&=&\sum_{(\boldsymbol{x},y)\in D^{(t)}} CE\left(\boldsymbol{q}\left(\boldsymbol{x}|\boldsymbol{\theta}\right),y\right),\nonumber\\
    L_{prior}\left(\boldsymbol{\theta}\right)&=&\sum_{k=1}^p F_k\left(\theta_k-\theta_k^{(t-1)}\right)^2,
\end{eqnarray}
where $\theta_k$ represents the $k$-th dimension of $\boldsymbol{\theta}$, $p$ represents the dimension of the parameter space, and $F_k$ is the $k$-th diagonal element of a Fisher information matrix. Here $L_{prior}$ constraints $\theta$ to be close to the Laplace approximation of the posterior Gaussian distribution with mean given by the previous model parameter $\boldsymbol{\theta}^{(t-1)}$ and a diagonal precision specified as the diagonal elements of the Fisher information matrix $F_k$'s. Given that the cross-entropy classification loss is proportional to the log-likelihood function, the diagonal elements of the Fisher information matrix can be computed as
\begin{equation}
    F_k=\left|\boldsymbol{D}^{(t)}\right|\cdot E\left[\frac{\partial^2}{\partial \theta_k^2}\,CE\left(\boldsymbol{q}(\boldsymbol{x}|\boldsymbol{\theta}),y\right)\big|\boldsymbol{\theta}\right],
\end{equation}
where $\left|\boldsymbol{D}^{(t)}\right|$ is the number of samples in the dataset $\boldsymbol{D}^{(t)}$.

The fundamental idea of the penalty term $L_{prior}$ is to \textbf{penalize large deviations from the previous classifier parameterized by $\boldsymbol{\theta}^{(t-1)}$, when training using $\boldsymbol{D}^{(t)}$, }so that the classifier with the updated model parameters can still perform well on samples from the old tasks. Specifically, the Fisher information $F$ contains information on the local curvature around the model parameter $\boldsymbol\theta$. Thus, discriminative penalties reflecting the local curvature are applied to the elements of $\boldsymbol\theta$ in such a way that a greater penalty is given to an element with a more rapid gradient. It implicates that the penalty term restricts important parameters to stay around $\boldsymbol\theta$, whereas allowing relatively non-informative parameters to move to learn new data. As a result, by minimizing $L_{prior}$, the model learns new data without spoiling what it has learnt from the previous tasks. Note that this does not create a significant computation load compared to the optimization without the penalty term as $F$ is readily computed from first-order derivatives. Adding this penalty term in turn allows updating the classifier while (i) maintaining previous accuracy of old defect types and (ii) reducing storage, computation and inspection needs as $\boldsymbol{D}^{(t)}$ only contains data from new defect types and a limited number of samples of the old defect types.

Since we place the same weight $\lambda_{ood}$ on the second and third terms in Eq.(\ref{eq:opt}), solving the optimization problem cannot control the trade-off between the false alarm and misdetection rates as was discussed in Section~\ref{ss:inspection}. Therefore, after updating $\boldsymbol{\theta}^{(t)}$ and $\boldsymbol{\phi}^{(t)}$, the threshold parameter $\tau^{(t)}$ should be tuned to achieve the desired new-defect detection performance. For example, we can specify the $\tau^{(t)}$ as a cut-off that guarantees an over $\eta\%$ true positive rate (for example, $\eta$ can be set as $80$), that is to compute
\begin{equation}
      \tau^{(t)}= \sup_\tau \left\{\tau:Pr\left[s\left(\boldsymbol{q}\left(\boldsymbol{x}|\boldsymbol{\theta}^{(t)}\right)|\boldsymbol{\phi}^{(t)}\right)
    \leq\tau\big|\boldsymbol{x}\in \boldsymbol{\widetilde{D}}^{(t)}\right]\geq \eta\%.\right\}
      \label{eq:tau}
\end{equation}
To this end, the parameters $\boldsymbol{\theta}^{(t)}$, $\boldsymbol{\phi}^{(t)}$ and $\tau^{(t)}$ are updated based on the inspected samples, which can be fed into the definitions of $f^{(t)}$ and $g^{(t)}$ to update the classifier and new-defect detector for future data batches. The algorithm is summarized in Algorithm~\ref{a:training}.

\begin{algorithm}[H]
    \SetAlgoLined
    \SetKwInOut{Input}{input}
    \SetKwInOut{Output}{output}
    
    \Input{labelled samples $\left\{\boldsymbol{D}^{(k)},k=0,...,t\right\}$, previous parameters $\boldsymbol{\theta}^{(t-1)},\boldsymbol{\phi}^{(t-1)},\tau^{(t-1)}$}
    
    \Init{}{Out-of-distribution dataset $\boldsymbol{\widetilde{D}}^{(t)}$}
    
    \textbf{Training in one epoch:}\\
      \For{$\left(\boldsymbol{X}_{in},\boldsymbol{Y}_{in}\right) \in \boldsymbol{D}^{(t)}$}{
        Randomly select $\left(\boldsymbol{X}_{out},\boldsymbol{Y}_{out}\right) \in \boldsymbol{\widetilde{D}}^{(t)}$\\
        \textbf{Forward Propagation:}\\
        $\boldsymbol{X}\gets \left[\boldsymbol{X}_{in},\boldsymbol{X}_{out}\right]$\\
        $\boldsymbol{Y}\gets \left[\boldsymbol{Y}_{in},\boldsymbol{Y}_{out}\right]$\\
        
        Evaluate $\boldsymbol{q}\left(\boldsymbol{X}|\boldsymbol{\theta}^{(t)}\right),s\left(\boldsymbol{q}\left(\boldsymbol{X}|\boldsymbol{\theta}^{(t)}\right)|\boldsymbol{\phi}^{(t)}\right)$\\

      \textbf{The value of Loss:}\\
      Evaluate $L\left(\boldsymbol{\theta}^{(t)},\boldsymbol{\phi}^{(t)},\tau^{(t-1)}\right)$ in Eq.(\ref{eq:opt})\\
      \textbf{Backward Propagation:}\\
      Evaluate $\frac{\partial L\left(\boldsymbol{\theta}^{(t)},\boldsymbol{\phi}^{(t)},\tau^{(t-1)}\right)}{\partial\left(\boldsymbol{\theta}^{(t)},\boldsymbol{\phi}^{(t)}\right)}$
      }
      Update $\tau$ in Eq.(\ref{eq:tau})\\
      \Output{updated model parameters $\boldsymbol{\theta}^{(t)},\boldsymbol{\phi}^{(t)},\tau^{(t)}$} 
      
    \caption{Continual Training}
    \label{a:training}
\end{algorithm}

\subsection{Model update}\label{ss:detector}
Lastly, the classifier $f^{(t)}$ and new-defect detector $g^{(t)}$ are updated based on the trained model parameters $\boldsymbol{\theta}^{(t)},\boldsymbol{\phi}^{(t)},\tau^{(t)}$ according to the definitions in Eq.(\ref{eq:f}) and Eq.(\ref{eq:g}), which is shown in Algorithm~\ref{a:update}. $g^{(t)}$ and $f^{(t)}$ will be used for identifying new defects and defect classification for future data batches, respectively. 

\begin{algorithm}[H]
    \SetAlgoLined
    \SetKwInOut{Input}{input}
    \SetKwInOut{Output}{output}
    \Input{updated parameters $\boldsymbol{\theta}^{(t)},\boldsymbol{\phi}^{(t)},\tau^{(t)}$}
        $f^{(t)}\left(\boldsymbol{x}\right)\gets \argmax\,\boldsymbol{q}\left(\boldsymbol{x}|\boldsymbol{\theta}^{(t)}\right),$\\
        $g^{(t)}\left(\boldsymbol{x}\right)\gets\mathbb{I}\left\{s\left(\boldsymbol{q}\left(\boldsymbol{x}|\boldsymbol{\theta}^{(t)}\right)\big|\boldsymbol{\phi}^{(t)}\right)\leq\tau^{(t)}\right\}.$\\
    \Output{updated classifier and detector $f^{(t)}$ and $g^{(t)}$}
    \caption{Model update}
    \label{a:update}
\end{algorithm}

\subsection{Practical considerations}\label{ss:practical}
In this subsection, we discuss some practical details when implementing the proposed approach to real engineering applications. The first practical issue is the choices of the auxiliary out-of-distribution dataset $\boldsymbol{\widetilde{D}}^{(t)}$. Adversarial out-of-distribution samples are commonly used in the out-of-distribution learning literature. For example, the adversarial samples can be generated via projected gradient descent (PGD, \cite{madry2017towards}) or fast gradient sign method (FGSM, \cite{goodfellow2014explaining}). However, these adversarial training techniques are designed to detect images under random disturbances rather than new types of defects with certain patterns. Under our setting, the principle is to require the samples in $\boldsymbol{\widetilde{D}}^{(t)}$ to resemble the potential new types of defects, such that a detector that is trained to separate $\boldsymbol{\widetilde{D}}^{(t)}$ from the old defects can also separate the future new types of defects from the old ones. In the engineering practice, it is recommended to leave out one old defect type as the out-of-distribution dataset, and train the new-defect detector to separate the other old types of defects from the left-out type of defect. Examples of selecting such an auxiliary out-of-distribution dataset will be given in the following two sections.

Second, it is recommended to utilize the limited storage space and improve the model performance via efficiently handling the previous and current data batches. Keeping the samples in the previous data batches can improve the updated classifier and new-defect detector's performance by preventing the classifier from forgetting the predictability of the old defects. After labelling the new defects in batch $t$, we suggest to draw a given number of samples in each of the old defect types within the memory budget. Then we run Algorithm~\ref{a:training} with a combination of the sampled old defects and the newly labelled defects. On the other hand, the classifier may have a poor prediction performance for the new defect types when the samples in the new defect types are not enough. To resolve this issue, we recommend to only update the classifier when the sample size of the new defect types exceeds a pre-specified threshold. When a particular defect type lacks training samples, the corresponding samples should be merged into the out-of-distribution dataset $\boldsymbol{\widetilde{D}}^{(t)}$ and re-train the new-defect detector. In this way, the detector can identify more samples in the target new defect type, send them to the inspection for labelling until a desired sample size of the new defect is achieved. 

In addition, the computational time and complexity of the proposed approach are dominated by the multiplication of the number of matrix operations during the back-propagation algorithm, the number of epochs, and the number of samples in the training dataset $\boldsymbol{D}^{(t)}$. The space complexity of the proposed approach is dominated by the multiplication of the number of samples in the training dataset $\boldsymbol{D}^{(t)}$ and the size of each sample. Comparing to the existing multitask learning approaches, the proposed method drops the dataset $\left\{\boldsymbol{D}^{(k)},k=1,\dots,t\right\}$ and only updates the model parameter based on the most recent dataset $\boldsymbol{D}^{(t)}$, which significantly improves the learning efficiency. Since both of the computation time and space complexities highly depend on $\left|\boldsymbol{D}^{(t)}\right|$, we recommend to firstly specify an upper limit of $\left|\boldsymbol{D}^{(t)}\right|$ to satisfy the constraints on computational resources, then balance the trade-off between the sample sizes of the old and new defects to achieve the desired classification accuracy.

Lastly, the hyper-parameters $\lambda_{ood}$ and $\lambda_{prior}$ in the loss function should be tuned before implementing the proposed algorithm to all the data batches. To reduce computational time, we suggest to tune the hyper-parameters based on the first several batches of data that contains new defect types. Specifically, the hyper-parameters should be selected to achieve the lowest classification error in the test dataset, while identifying over a pre-specified percentage (e.g., $50\%$) of the new types of defects for new data batches. Detailed discussion on the hyper-parameter tuning procedure can be found in Appendix B.

\section{Simulation study}\label{s:simulation}
\subsection{Performance of the trained detector and classifier}\label{ss:simu_result}
\begin{figure}
    \centering
    \includegraphics[height=2.5in]{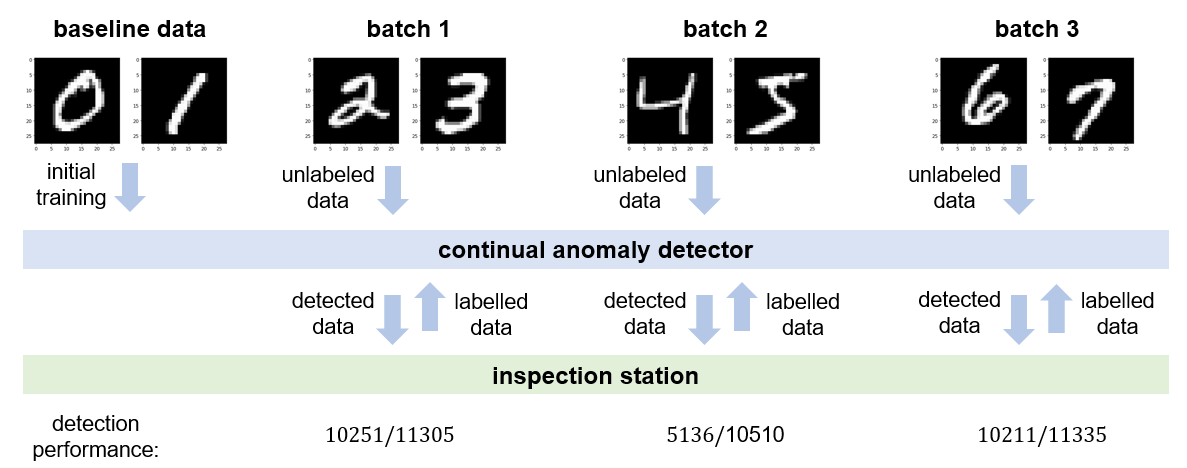}
    \caption{\it{Illustration of continual anomaly detection in the MNIST dataset. The top panel shows exemplar images with different class labels in each batch. The bottom panel shows the ratio of detected samples (the values before ``$/$'') among all the samples in the new classes (the values after ``$/$'').}}
    \label{fig:mnist_flow}
\end{figure}

In this section, we demonstrate the effectiveness of the proposed approach by a simulation study on the Modified National Institute of Standards and Technology (MNIST) dataset \citep{deng2012mnist}. The objective is to train a classifier that learns the digit labels from images of handwritten digits. To fit the MNIST dataset into the adaptive defect classification situation, we assume that a classifier is pre-trained based on the images with digits $\{0,1\}$, and then updated based on the images with other digits in three data batches sequentially. Batches $1$, $2$ and $3$ only include digits $\{2,3\}$, $\{4,5\}$, and $\{6,7\}$, respectively. In each class, we extract $4-$dimensional features from the flattened $784\times 1$ images via a trained deep auto-encoder, and then divide $80\%$ of the samples into the training dataset, and the rest into the test dataset using ``SPlit'' \citep{vakayil2021package, joseph2021split}. We leave the images with digits $\{8,9\}$ as a potential choice of the out-of-distribution dataset. As shown in Fig.~\ref{fig:mnist_flow}, we presume that the images are not labelled at each batch, and apply the proposed approach to detect images from new classes and label these images in the inspection station. Afterwards, the labelled images are fed back to the continual learner for model updating.

Given the limited storage space, we keep $3,000$ images in each old class when processing the current data batch. We select the benchmark classifier Residual Network (ResNet \citep{he2016deep}) with two blocks, each of which is consisted of two convolution layers with kernel size $3$. We tune the hyper-parameters following the procedure in Section~\ref{ss:practical}. The resultant $\lambda_{prior}=\lambda_{ood}=1$. We compare the new class detector's performance based on the ODIN and Mahalanobis-distance-based scores. We set $\eta=80$ in Eq.(8) to allow for $20\%$ false positive rate in detecting new classes. We train the baseline model with digits $\{0, 1\}$, train the out-of-distribution detection with digits $\{8, 9\}$, and test the detection performance in the three sets of new digits $\{2, 3\}$, $\{4, 5\}$ and $\{6, 7\}$. The detector based on the ODIN score detects $84.25\%$, $86.85\%$ and $80.40\%$ samples in the three sets of new digits, respectively. The detector based on the Mahalanobis-distance-based score detects $88.95\%$, $88.92\%$, and $88.14\%$, respectively. We select the ODIN score in the following analysis because Mahalanobis-distance-based score requires more computational time for a complex deep neural network although the Mahalanobis-distance-based score show a slight better detection performance than the ODIN score.

\begin{figure}
    \centering
    \includegraphics[height=2.7in]{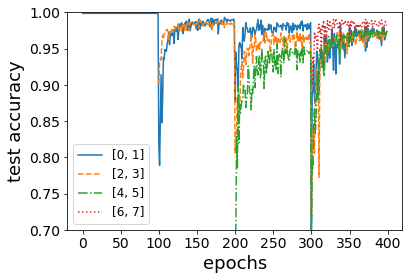}
    \caption{\it{Prediction accuracy of the classifier in the test dataset. The solid line depicts the test accuracy of class $\{0,1\}$ at different epochs during baseline training and the three batches. The dashed line represents the test accuracy of class $\{2,3\}$ at different epochs in the three batches. The dot-dashed line shows the test accuracy of class $\{4,5\}$ at different epochs in batches $2$ and $3$. The dotted line represents the test accuracy of class $\{6,7\}$ at different epochs in batch $3$. Each task is trained for $100$ epochs. The test accuracy of the old tasks drops to relatively lower values at the initial $20$ to $30$ epochs after each $100$ epochs. This is because the corresponding model parameter deviates from that in the previous task while still not reaches the minimal point of the proposed loss function. However, after training for enough epochs, the test accuracy of both the old and new tasks increase to above $0.97$, thus demonstrating the effectiveness of the proposed method.}}
    \label{fig:mnist_result}
\end{figure}

The proposed approach is applied to the MNIST dataset under the above setting. The classifier is trained $100$ epochs for each batch of data. The performance is evaluated based on the false negative rates of the detector $g^{(t)}$ and the prediction accuracy of the classifier $f^{(t)}$ in the test dataset. The true negative rates in all the three batches are displayed at the lower panel of Fig.~\ref{fig:mnist_flow}, which implies that the trained detector can detect more than $48.8\%$ of the samples in the new classes, guaranteeing that more than $2,500$ samples of each new class are used for model update. Fig.~\ref{fig:mnist_result} shows the prediction accuracy of different classes in different batches. After $100$ training epochs, the prediction accuracy is above $97\%$ for all the classes, demonstrating that the proposed continual learner identifies the new class labels without forgetting the old ones. The prediction accuracy results also validate our arguments in Section~\ref{ss:inspection} that the new-defect detector is not required to have a near-perfect performance in catching all the samples of new classes. Instead, we can train a good classifier with enough samples from new classes, classify the remaining unlabelled samples in the old data batches, and then add the classified samples for the future data batches. 

\subsection{Choice of auxiliary out-of-distribution dataset}
\begin{figure}
    \centering
    \includegraphics[height=1.9in]{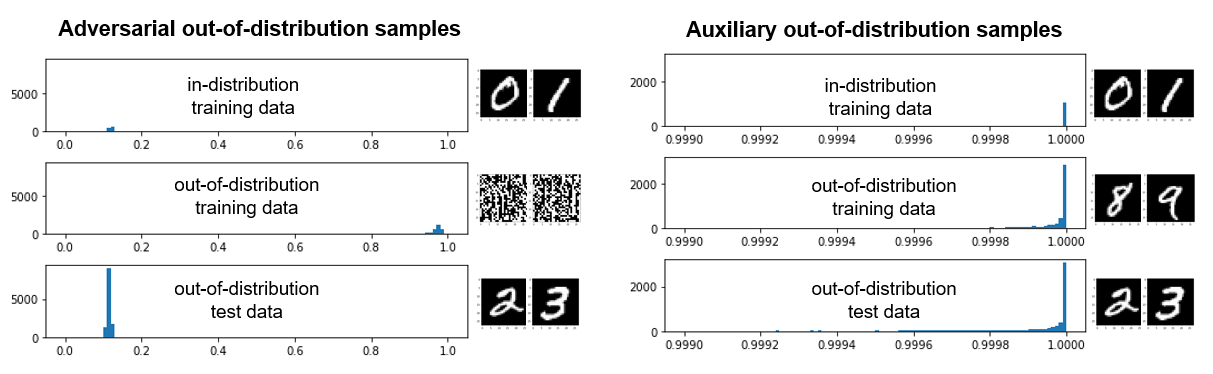}
    \caption{\it{Comparisons between the adversarial out-of-distribution data and the actual auxiliary out-of-distribution data. The three rows depict the histograms of the score function $s$ for the in-distribution training data, the out-of-distribution training data, and the out-of-distribution test data, respectively. The left panel indicates that the adversarial out-of-distribution samples are easier separable from the in-distribution samples than the actual out-of-distribution test data, making the initially trained detector fail to detect out-of-distribution samples in the future batches. The right panel implies that an auxiliary out-of-distribution data that has a similar writing pattern as the other out-of-distribution samples improves the detector performance.}}
    \label{fig:mnist_adv}
\end{figure}

There are multiple choices of auxiliary out-of-distribution samples. For illustration purpose, we compare the detection performance of two detectors ($g$), one is trained using the adversarial samples generated from projected gradient descent (PGD, \citep{madry2017towards}) as the out-of-distribution dataset ($\boldsymbol{\widetilde{D}}^{(t)}$), the other is trained using the images with labels in $\{8,9\}$ as the out-of-distribution dataset. The comparison results are illustrated in Fig.~\ref{fig:mnist_adv} with a baseline model trained with samples in class $\{0,1\}$. On the left panel of Fig.~\ref{fig:mnist_adv}, the generated adversarial samples do not resemble the actual out-of-distribution samples, which results in a detector that cannot identify out-of-distribution samples with labels $\{2,3\}$. On the other hand, as the samples in classes $\{8,9\}$ share the similar out-of-distribution pattern as the samples in classes $\{2,3\}$ (shown on the right panel of Fig.~\ref{fig:mnist_adv}), a detector that separates the auxiliary samples in $\{8,9\}$ from $\{0,1\}$ can also separate the actual out-of-distribution samples in $\{2,3\}$ from $\{0,1\}$.

\section{Case study}\label{s:case}
\begin{figure}
    \centering
    \includegraphics[height=3.5in]{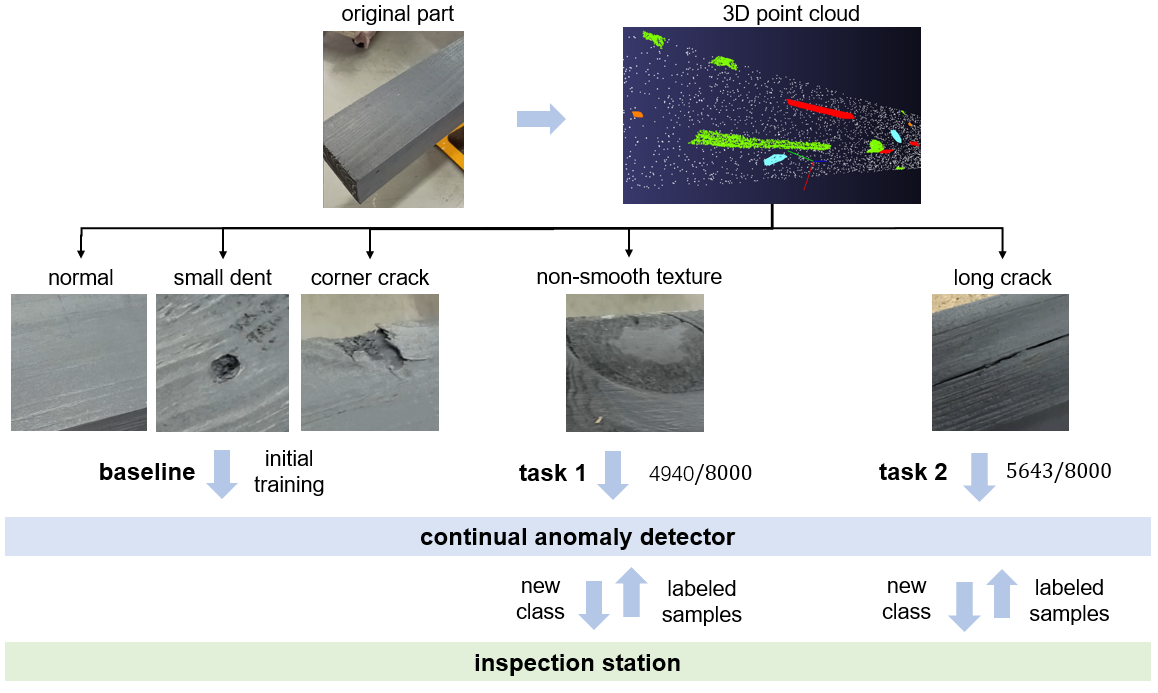}
    \caption{\it{Illustration of continual anomaly detection in the 3D point cloud dataset. The top panel illustrates a segment of the original part, and an exemplar segment of the 3D point cloud data. The light points represent the center points of the normal class. The dark points represent the center points of different defects. The middle panel shows a normal surface and four types of defects. The bottom panel shows the ratio of detected samples (the values before ``$/$'') among all the samples in the new classes (the values after ``$/$'').}}
    \label{fig:3dpt_flow}
\end{figure}

In this section, we apply the proposed method to a 3D point cloud dataset generated by scanning a long-cuboid-shaped wood part with a 3D laser scanner. There exists five major types of defects on the four side surfaces of the wood part: small dent, corner crack, big dent, long crack, and non-smooth texture. We manually highlight the defective areas after importing the 3D point clouds into the software ``MeshLab'' (shown in the top panel of Fig.~\ref{fig:3dpt_flow}). We adopt the data-augmentation strategy to increase the sample size, which is implemented in the following three steps. (i) For each class, including the four types of defects and the normal surface, we sample $10,000$ points from the highlighted areas in each class as the center points. (ii) Given each center point, we sample $300$ points that come from the same highlighted area and lie within $10mm$ from the center point. The distance threshold $10mm$ is specified based on our prior knowledge on the sizes of defects. We treat the $300$ points as one sample, which is stored as a $300\times 3$ 3D point cloud matrix with a class label. (iii) We treat each 3D point cloud matrix as an image to employ the existing image classification techniques. Due to the fact that different permutations of the rows of a 3D point cloud matrix represent the same 3D object, we sequentially sort each 3D point cloud matrix based on the third and second columns (representing the Z and Y coordinates in the Cartesian coordinate system, respectively). In this way, our samples are invariant to the permutation variation, which is a key challenge in the area of 3D point cloud data analytic. To this end, we have $10,000$ samples for each data class. Each sample contains a $300\times 3$ image and a class label. Similar to the analysis in Section 4, in each class, we extract $10-$dimensional features from the flattened $900\times 1$ images, and then divide the samples into a training dataset of $8,000$ samples and a test dataset of $2,000$ samples using ``SPlit'' \citep{vakayil2021package, joseph2021split}.

To simulate the scenario where high-volume point clouds come from multiple data batches, we assume that samples from the normal surface, small dent, and corner crack classes are available in the initial batch. Then the samples from the non-smooth texture and long crack classes come in two batches sequentially. In these two batches, we presume the sample labels are not known until sending to the inspection station. For the reasons indicated in Section~\ref{s:simulation}, we use the samples from the big dent class as auxiliary out-of-distribution samples for training the new-defect detector. The proposed approach is then applied for adaptive defect classification of the 3D point cloud data batches. Specifically, we train the classifier as a neural network for image classification, which includes two convolution layers and three fully connected layers. Each convolution layer is consisted of a 2D convolution with kernel size $3$ and a maxpooling layer of size $3$. The first convolution layer outputs $6$ features, the second convolution layer outputs $16$ features. The output dimensions of the three fully connected layers are $120$, $84$, and $7$, respectively. In each batch, the neural network is trained for $200$ epochs with a learning rate $0.0001$. Based on the test performance, we set $\lambda_{ewc}=\lambda_{ood}=0.1$. To guarantee the classification performance, we keep $3,000$ out of $8,000$ training samples from the previous batches when learning the new batch. 

The performance of the proposed approach is evaluated based on three criteria - (i) the new-defect detector should detect enough new defects for model update, (ii) the classifier should separate defects from normal surfaces, and (iii) the classifier should also identify the exact defect types. For criterion (i), we check the proportions of detected new types of defects in the training dataset of each data batch, which are illustrated in the bottom panel of Fig~\ref{fig:3dpt_flow}. In the first batch, the new-defect detector identifies $4,940$ out of $8,000$ non-smooth texture samples as the new type of defects that have not been seen in the baseline dataset. In the second batch, the new-defect detector screens $5,643$ long crack samples from the total $8,000$ training samples. As was indicated in Section~\ref{ss:inspection}, these proportions are not necessarily to be close to $1$ once the sample size of the newly detected defects is enough for updating the model.

For criterion (ii), we check the number of non-defective test samples that are incorrectly classified as defects, as well as the number of defective test samples that are incorrectly recognized as normal. In both of the two data batches, the number of mis-classified cases is $4$ among $2000$ normal samples in the test dataset. In data batch $2$, $1$ sample of long crack are incorrectly recognized as normal surfaces among a total of $8,000$ defect samples. These results indicate the trained classifier is powerful in detecting surface defects.

\begin{table}[t]
\caption{Confusion matrix of the classifier}
\begin{center}
\label{t:conf_mat}
\begin{tabular}{c c c c c c}
& & \\ 
\hline
Class & Normal & Small Dent &  Corner Crack & Long Crack & Texture\\
 & (true) & (true) & (true) & (true) & (true)\\
\hline
Normal (predict) & 1996 & 0 & 0 & 1 & 0 \\
Small Dent (predict) & 4 & 1913 & 69 & 29 & 6 \\
Corner Crack (predict) & 0 & 11 & 1811 & 246 & 5 \\
Long Crack (predict) & 0 & 46 & 86 & 1690 & 74 \\
Texture (predict) & 0 & 30 & 34 & 34 & 1915 \\
\hline
\end{tabular}
\end{center}
\end{table}

\begin{figure}
    \centering
    \includegraphics[height=4.5in]{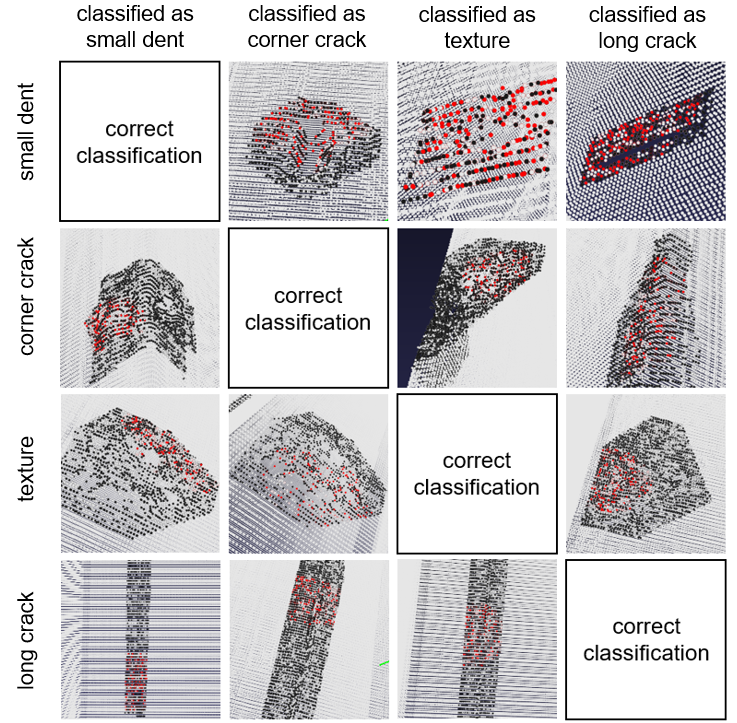}
    \caption{\it{Exemplar mis-classified samples. Each row represents a true defect type. Each column represents a defect type based on the trained classifier. Each off-diagonal figure represents an example of mis-classification. The red dots highlight the sets of points in the samples. The black dots illustrate the actual defective area. The grey dots depict the normal surface.}}
    \label{fig:3dpt_class}
\end{figure}

For criterion (iii), the classification performance is illustrated in Table~\ref{t:conf_mat}, which is the confusion matrix of the classifier in the test dataset after training all the data batches. The proposed method has an overall test accuracy of $\frac{1996+1913+1811+1690+1915}{2000+2000+2000+2000+2000}=93.25\%$. Note here that the accuracy of catching a defect (i.e. Normal vs non-normal) is \textbf{99.99\%}. Yet the ability of the model to separate all the classes is $93.25\%$. Comparing to the number of mis-classified cases in criterion (ii), the defect classification has a relatively higher classification error, which mainly arises from the strategy of generating the 3D point cloud samples. Fig.~\ref{fig:3dpt_class} shows exemplar mis-classified samples under different combinations of the true defect types and the predicted defect types. The subplot at block $(1,2)$ shows a sample drawn from a small dent area. Since the center of the dent has an abnormal shape, the sample is recognized as a corner crack. The subplot at block $(1,3)$ illustrates a sample from a small dent area whose depth is not large. In this case, the classifier classifies the sample as a non-smooth texture. The subplot at block $(1,4)$ illustrates an oval-shaped small dent. The sample is classified as a long crack since the width of the dent is much smaller than the length of the dent. In the subplot at blocks $(2, 1)$ and $(2, 3)$, since the samples are drawn from one surface on the corner crack, they are identified as a small dent and a non-smooth texture, respectively. The sample at block $(2,4)$ resembles a long crack, giving rise to a mis-classification error. In the third row, a non-smooth texture could be classified as other types of defects when the surface variation becomes larger. In the last row, when the points are only sampled from a small segment of the long crack, they are likely to be classified as other types of defects, which gives majority of the mis-classification cases in Table~\ref{t:conf_mat}. In all, due to the vague boundary among different types of defects and the variation in sampling the points from point clouds, the overall classification performance is not as good as the defect detection performance. However, the mis-classification examples in Fig.~\ref{fig:3dpt_class} still implies that the classifier produces reasonable results matching with the human knowledge on the defect shapes.

\section{Conclusion}\label{s:conclusion}
A new adaptive defect classification framework is proposed for high-volume independent data batches. The integration of a continual learner and an out-of-distribution detector enables an effective inspection of unlabelled samples and updating the model upon the arrival of new defect data. The simulation and case study have shown the effectiveness of the proposed method.

The quality inspection in the case study is to check surface defects of a wood log that has four flat surfaces. This allows us to transform the 3D point clouds into surface defects' images and train a image classifier. It should be clarified that the proposed statistical framework can be incorporated with any commonly-used classifier including for 3D point clouds. If defect detection is made on complex 3D objects, we recommend to apply a 3D point cloud classification tool as a classifier (for example, PointNet \citep{qi2017pointnet}).

The proposed statistical framework is tailored for the discretized data batches. However, it can be extended to the cases where data batches are generated from continuous manufacturing processes. Take the roll-to-roll manufacturing as an example, such as paper or fabric manufacturing process, we can divide the part surface into equally large non-overlapped grids, and then continuously inspect grids via a laser scanner. Our proposed approach can be applied to the scanned 3D point cloud data in each grid for defect classification during continuous production. The model is then updated when enough samples in the new defect types are collected. It is worth noting that when continuous monitoring data are collected, future works need to address the auto-correlations among the samples within each grid.

We end by noting that although a practical approach for selecting the auxiliary out-of-distribution dataset has been provided in Section~\ref{ss:practical}, it relies on whether we have additional old defect types to leave out, which may not be available in real applications. As a future work, a systematic approach should be developed to automatically generate out-of-distribution samples that resemble the patterns of samples in the new defect types. For example, a valid out-of-distribution sample for the MNIST dataset should contain some random handwritten scratches instead of random mosaics.

\bibliography{Raed-ref}
\end{document}


\maketitle

\section*{Appendix A: definition of the new-defect detector using ODIN score}
When the ODIN score is employed, the score function is defined as
\begin{equation}
    s\left(x|\phi^{(t)}\right)\equiv\max\,\tilde{q}_i\left(\tilde{x}| \phi^{(t)}\right),
    \label{eq:odin}
\end{equation}
where $\phi=\left\{T,\epsilon\right\}$ collects the temperature and disturbance parameters, $\tilde{q}_i\left(\cdot|\phi^{(t)}\right)$ represents the predicted probabilities after temperature scaling, and $\tilde{x}$ represents the input under perturbation. Specifically, the temperature scaling function $\tilde{q}_i$ is defined as
\begin{equation}
    \tilde{q}_i\left(x|\phi^{(t)}\right)\equiv \frac{\exp\left(q_i\left(x|\theta^{(t)}\right)/T\right)}{\sum_{j=1}^{\left|\mathcal{Y}^{(t)}\right|}\exp\left(q_j\left(x|\theta^{(t)}\right)/T\right)}.
    \label{eq:temperature}
\end{equation}
The perturbed input can be computed as
\begin{equation}
    \tilde{x}=x-\epsilon \text{sign}\left(-\nabla_x \log \tilde{q}_{f^{(t)}(x)}\left(x|\phi^{(t)}\right)\right),
    \label{eq:perturbation}
\end{equation}
where $\text{sign}(\cdot)$ is the sign function, and $\nabla_x$ is the gradient operator with respect to $x$. The score function in Eq.(\ref{eq:odin}) maps an input $x$ into a real-valued score $s$.

\section*{Appendix B: choice of the hyper-parameters}
There are three key hyper-parameters to be tuned in the proposed method - $\tau^{(t)}$, $\lambda_{ood}$ and $\lambda_{prior}$. The threshold $\tau^{(t)}$ is tuned to guarantee an over $\eta\%$ true positive rate as was shown in Eq.(8). Here we only discuss the choices of $\lambda_{ood}$ and $\lambda_{prior}$.

The hyper-parameter $\lambda_{ood}$ affects whether the score function can separate new defects from old ones. Since the score function is established on the trained classifier (as was shown in Eq.(4)), an over-parameterized deep neural network can be trained to simultaneously guarantee a low classification loss and an effective separation between old and new defects. In other words, the loss term $L_{cont}(\theta)$ and the penalty term in Eq.(5) do not contradict each other, and thus the classifier's performance should not be very sensitive to the choice of $\lambda_{ood}$. 

To validate this intuition, we fix $\lambda_{prior}=1$, and test the proposed method in the MNIST dataset by setting the value of $\log_{10}\left(\lambda_{ood}\right)$ as $\left\{-2, -1, 0, 1, 2, 3\right\}$, respectively. The left panel of Fig.~\ref{fig:lambda_ood} shows the test accuracy of the old class ${0,1}$ and the new classes ${2, 3}$ and ${4, 5}$ under different choices of $\lambda_{ood}$, which indicates that the proposed method is not sensitive to the values of $\lambda_{ood}$.
\begin{figure}
    \centering
    \includegraphics[height=2in]{rebuttal/lambda_ood.png}
    \includegraphics[height=2in]{rebuttal/lambda_prior.png}
    \caption{\it{Illustration of the prediction accuracy in the test dataset under different $\log_{10}\lambda_{ood}$ (left panel) and $\log_{10}\lambda_{prior}$ (right panel). The solid line depicts the test accuracy of class $\{0,1\}$ at different epochs during baseline training and when the three data batches arrive. The dashed line represents the test accuracy of class $\{2,3\}$ at different epochs when the three data batches arrive. The dot-dashed line shows the test accuracy of class $\{4,5\}$ at different epochs when the last two data batches arrive.}}
    \label{fig:lambda_ood}
\end{figure}

The hyper-parameter $\lambda_{prior}$ encourages the optimizer to update the model parameter of the classifier around the vicinity of the model parameter, $\boldsymbol\theta^{(t-1)}$ in the previous task, which prevents the mis-classification of the samples in the old tasks. Intuitively, training with a small $\lambda_{prior}$ results in a classifier with poor prediction performance on the samples in the old tasks. A large $\lambda_{prior}$ ties the model parameter around the vicinity of $\boldsymbol\theta^{(t-1)}$, and thus results in a classifier that cannot learn the new tasks. In section 3.5, we suggest to draw a given number of samples in each of the old defect types within the memory budget, such that the classifier can still predict both the old and new tasks when the model parameter is not close to $\boldsymbol\theta^{(t-1)}$. This treatment introduces the robustness of the proposed method with respect to the choice of $\lambda_{ood}$.

To validate the arguments above, we fixed $\lambda_{ood}=1$, and varied the values of $\log\left(\lambda_{prior}\right)$ in the set $\left\{-2, -1, 0, 1, 2, 3\right\}$. The right panel of Fig.~\ref{fig:lambda_ood} shows the test accuracy of the old class ${0,1}$ and the new classes ${2, 3}$ and ${4, 5}$ under different values of $\lambda_{prior}$. The figure illustrates that the method performance is poor when $\lambda_{prior}$ is greater than $100$. Given that the model performance is not sensitive to $\lambda_{ood}$, we used the widely-used grid-search method to determine the univariate hyper-parameter, $\lambda_{prior}$. We selected $\lambda_{prior}=1$ because the algorithm under $\lambda_{prior}=1$ returns the highest average prediction accuracy among all the classes, which is $0.9572$. 

Furthermore, we compared the model performance under different numbers of old samples that are brought back to the training dataset. The comparison is visualized in Fig.~\ref{fig:old_sample_size}, where the averaged test accuracy for the three tasks increases with the sample size of old samples. Although more samples in the previous batch can improve the model performance, we would recommend stop introducing old samples whenever the test accuracy satisfies practitioner's expectation or the number of old samples exceed the memory budget.
\begin{figure}
    \centering
    \includegraphics[height=3in]{rebuttal/old_sample_size.png}
    \caption{\it{Illustration of the prediction accuracy in the test dataset under different sample size in the previous batch. The solid line depicts the test accuracy of class $\{0,1\}$ at different epochs during baseline training and when the three data batches arrive. The dashed line represents the test accuracy of class $\{2,3\}$ at different epochs when the three data batches arrive. The dot-dashed line shows the test accuracy of class $\{4,5\}$ at different epochs when the last two data batches arrive.}}
    \label{fig:old_sample_size}
\end{figure}